\def\year{2021}\relax
\documentclass[letterpaper]{article} 
\usepackage{aaai21}  
\usepackage{times}  
\usepackage{helvet} 
\usepackage{courier}  
\usepackage[hyphens]{url}  
\usepackage{graphicx} 
\usepackage{natbib}  
\usepackage{caption} 
\usepackage{graphicx}
\usepackage{booktabs, tabularx}
\usepackage{subfigure}
\usepackage{url}
\usepackage{amsmath, amssymb}
\newcommand\numberthis{\addtocounter{equation}{1}\tag{\theequation}}
\usepackage[ruled,vlined]{algorithm2e}
\usepackage{mathrsfs}

\usepackage[switch]{lineno} %

\title{Game of Gradients: Mitigating Irrelevant Clients in Federated Learning}
\author {
        Lokesh Nagalapatti\thanks{Work done while at IBM Research-India},\textsuperscript{\rm 1} 
        Ramasuri Narayanam\textsuperscript{\rm 2} \\
}
\affiliations {
    \textsuperscript{\rm 1} IIT Bombay, Mumbai India \\
    \textsuperscript{\rm 2} IBM Research - India \\
    nlokeshiisc@gmail.com,
    ramasurn@in.ibm.com
}

\begin{document}

\maketitle

\begin{abstract}
The paradigm of {\em Federated learning (FL)} deals with multiple clients participating in collaborative training of a machine learning model under the orchestration of a central server. In this setup, each client’s data is private to itself and is not transferable to any other client or the server. Though FL paradigm has received significant interest recently from the research community, the problem of selecting relevant clients w.r.t. the central server's learning objective is under-explored. We refer to these problems as Federated Relevant Client Selection (FRCS). Because the server doesn't have explicit control over the nature of data possessed by each client, the problem of selecting relevant clients is significantly complex in FL settings.
In this paper, we resolve important and related FRCS problems viz., selecting clients with relevant data, detecting clients that possess data relevant to a particular target label, and rectifying corrupted data samples of individual clients. We follow a principled approach to address the above FRCS problems and develop a new federated learning method using the Shapley value concept from cooperative game theory. Towards this end, we propose a cooperative game involving the gradients shared by the clients. Using this game, we compute Shapley values of clients and then present {\em Shapley value based Federated Averaging (S-FedAvg)} algorithm that empowers the server to select relevant clients with high probability. {\em S-FedAvg} turns out to be critical in designing specific algorithms to address the FRCS problems. We finally conduct a thorough empirical analysis on image classification and speech recognition tasks to show the superior performance of S-FedAvg than the baselines in the context of supervised federated learning settings.
\end{abstract}

\section*{Introduction}
\label{sec:intro}
Federated Learning (FL) \cite{mcmahan:2017} is a machine learning paradigm where the objective is to collectively train a high-quality central model while the training data remains distributed over a number of clients. The data that is owned by each client is private to itself and not transferable to other clients or the server. Hence, each client participates in FL by sharing updates derived from data rather than sharing the data itself. One popular algorithm that is used in FL settings is called federated averaging algorithm (FedAvg) \cite{mcmahan:2017}. This framework is different from distributed optimization setting \cite{smith-jmlr-2018,recht-nips-2011} primarily in two aspects: (a) {\em Non-IID Data}: Any particular client's local dataset will not be representative of the population distribution; (b) {\em Unbalanced Data}: The size of private data owned by individual clients could vary significantly. FL setting assumes a synchronous update of the central model and the following steps proceed in each round of the learning process \cite{mcmahan:2017}. There is a fixed set of $K$ clients and each client owns a fixed local dataset. In each round, the central server randomly selects a fraction of clients. Each selected client downloads the state of current central model and then performs local computation based on its local dataset. Finally, it sends the model updates to the server. The server then aggregates the updates received from each selected client and applies them to its global model. The state of the central model is thus updated and this process repeats until the central model converges. Note that only a fraction of clients is selected in each round as adding more clients would lead to diminishing returns beyond a certain point  \cite{mcmahan:2017}. Though initially the emphasis was on mobile-centric FL applications involving thousands of clients, recently there is significant interest in enterprise driven FL applications that involve only a few tens of clients \cite{dinesh-verma-2019}.
Though FL paradigm has received significant interest recently from the research community, the problem of assessing quality of the individual client's private data with respect to the central server's learning objective is still under-explored and we refer to this as {\em Federated Relevant Client Selection (FRCS) problem}. This is an important problem as selecting clients with irrelevant or corrupted data could annihilate the stability of the central model. We call a client as {\em irrelevant} if the updates shared by it adversely affects the performance of the central model. Next, we motivate the need for studying FRCS problems.



\begin{figure}[!hbtp]
    \centering
    \includegraphics[width=0.95\linewidth]{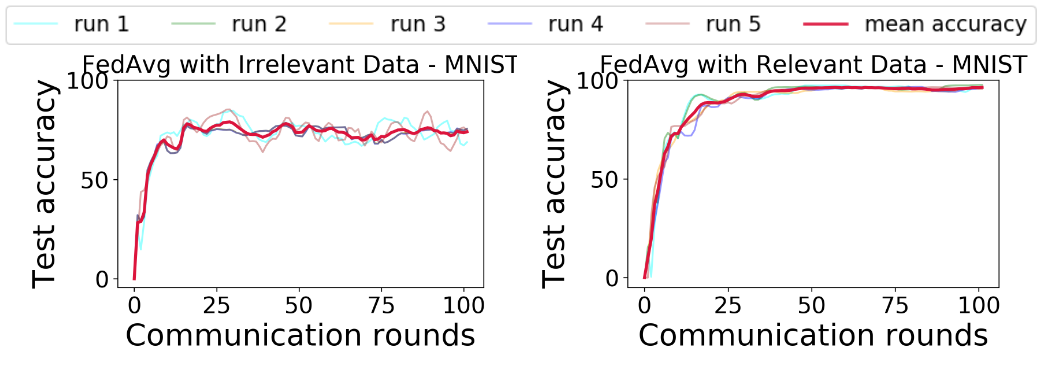}
\caption{Performance of FedAvg algorithm applies to (a) $10$ relevant clients (b) $6$ relevant and $4$ irrelevant clients.}
\label{fig:motivation}
\end{figure}



{\bf Motivation:} To motivate the impact of clients' local data quality on the central model, we apply standard FedAvg algorithm to two cases -- (a) where all clients possess relevant data (b) where some clients possess irrelevant data. The central server tries to build a three layered neural network that classifies even labeled hand written digits i.e., \{$0,2,4,6,8$\}. In this setting, all the data with odd labeled digits becomes {\em irrelevant} w.r.t. the central server's objective. We work with MNIST dataset and distribute the training data to $10$ clients. In case (a), we distribute even-labeled data to each of the $10$ clients and in case (b) we distribute even labeled data to $6$ clients and irrelevant odd labelled data to remaining $4$ clients. To simulate irrelevance, we work with open-set label noise \cite{openset_noise} strategy, wherein we randomly flip each odd label of the $4$ irrelevant clients to one of the even labels.
Figures \ref{fig:motivation}$(a)$ and \ref{fig:motivation}$(b)$ show the accuracy of the model on the test data across $100$ communication rounds for case (a) and case (b) respectively. The graph curves {\em run $1$} to {\em run $5$} show accuracy in $5$ independent runs and the dark red line corresponds to average of the accuracies in these runs. As expected, we can see that the model convergence is smooth in case (a) when all the clients are relevant.
In case (b), even though the relevant clients are in a majority, the updates incorporated from a few irrelevant clients affect the overall stability of the model. 
This experiment demonstrates the need to address the limitation of existing federated learning methods to deal with clients having irrelevant or corrupted data. 

{\bf Contributions of Our Work:} Here we briefly summarize the primary contribution of this work: 
\begin{itemize}
    \item We model FL as a cooperative game with updates received from clients as players and performance of the model on the server's validation dataset as the characteristic function. We compute Shapley value for each client and use it to assess its relevance w.r.t. the server's learning objective.
    \item Using the Shapley values, we propose an algorithm called Shapley values based federated averaging (S-FedAvg) that empowers the server to select relevant clients with high probability.
    \item Using S-FedAvg, we propose techniques to solve two FRCS problems: (a) Class-specific best client selection: detect clients that have good quality data corresponding to a particular target label; and (b) Data Label Standardization: detect and correct corrupted data samples of clients.
    \item Extensive experimentation that shows the efficacy of our solution approach.
\end{itemize}





\section*{Related Work}
\label{sec:related_work}
We review the literature related to our work by dividing them into three categories. 

\textbf{Federated Learning:} We refer the reader to \cite{yang-survey-2019} for a detailed discussion on FL and its challenges. Research in FL has focused on several aspects. On privacy aspect, many solutions are proposed to provide strong privacy guarantees that protect the confidentiality of the clients' data \cite{privacy1, privacy2, privacy3}. Privacy guarantees are important because in FL setting, clients usually share the gradients derived from their local data. Recently, \cite{deepleekage} has shown a mechanism to recover the input data completely given its gradient. 
On efficiency aspect, research has been done to improve the communication efficiency \cite{communication_efficiency}, building platforms that can robustly facilitate  participation of thousands of clients \cite{federated_scale}, etc. Because of the Non-IID and unbalanced nature of data distribution in FL, optimization algorithms like SGD may produce sub optimal results and lack theoretical understandings \cite{sgd_fed_convergence, li2019convergence}. This is primarily because the stochastic gradient derived from non-IID data need not necessarily form an unbiased estimate of gradient on full data. \cite{federatedNonIID} showed a method to overcome the non-IID issue with an assumption that 5\% of test data is made publicly available by the server.

\textbf{Irrelevant Clients in FL: }A client is called {\em relevant} if the data it holds, and hence the updates it shares, helps to improve the performance of the central model. We call a client as {\em irrelevant} if the updates shared by it turn detrimental to the central model. A client's data can be irrelevant due to a number of reasons like label-noise, outliers, data not following the target distribution, noise in the attributes/predictor variables, etc.   Many techniques are proposed to tackle the different kinds of data irrelevancies; some of which include \cite{label_noise_approach_1, federatedNonIID, outlier_detection}. A new line of research includes works that show methods to select a subset of training data that aligns well with the model's objective \cite{datashapley, choicenet}. These methods have shown that sometimes ignoring a portion of irrelevant training data helps boost the model performance significantly. However all the above mentioned works make an assumption that all the training data is available at one place and hence is challenging to extend them to FL settings.

\textbf{Clients Selection in FL: }More recently, research has started to focus on detecting irrelevant clients that participate in FL. The updates shared by irrelevant clients are either ignored or suppressed when they are aggregated by the server to be incorporated into the central model. \cite{client_selection_2} show that the FedAvg algorithm is not tolerant to the presence of adversaries and propose an aggregation method called Krum that incorporate updates only from benign clients. \cite{client_selection_1} propose an aggregation operator called IOWA-DQ that down weights the updates from irrelevant/adversarial clients. \cite{client_selection_3} model FL as a stackelberg game that incentivices the clients participating in it. Server allocates a reward to encourage client participation based on the amount of contribution made by it. Each client then tries to maximize the reward subject to cost constraints like compute, data procurement costs etc. The paper has further shown the existence of Nash equilibrium when every data point at each client is IID which is not a realistic assumption in FL. \cite{client_selection_4} propose an incentive mechanism that measures the contribution of each client by means of Shapley values. To compute the contribution made by each client, this method needs updates to be shared from all the clients at each round of FL. While this approach focuses on how to measure the contribution of each client, our proposed approach besides using Shapley values to model the contribution of clients, it also dynamically selects clients that are relevant and likely to contribute more in the next round with high probability. \cite{client_selection_5} propose a scheme to incentivize FL by means of contracts. In this approach, server devises a contract and advertises it to the clients. Each client examines the contract and decides to participate if it finds the contract satisfactory. However, unlike our proposed framework, this approach makes a sort of unrealistic assumptions that the server has knowledge about both (a) client's data quality and (b) client's compute resources so that server can devise high paying contracts to relevant clients only.






\section*{Problem Statement}

We first formalize the standard federated learning framework \cite{mcmahan:2017}. It consists of a set $S = \{1,2, \ldots, K\}$ of $K$ clients/nodes wherein each client $k$ has access to data $\mathbb{D}_k$ such that $|\mathbb{D}_k| = n_k$. Let $n$ be total number of training samples, $n = \sum_{k \in K}n_k$. The data $\mathbb{D}_k$ is locally accessible to node $k$ and is not shared to any other node as well as to the central server, $C$. The objective of the server $C$ is to train a classifier $f(\theta)$, where $\theta$ is a global parameter vector obtained by distributed training and aggregation over $K$ clients with the hope that it generalizes on the test dataset $D_{test}$. In federated learning setting, the objective of the central server takes the following form:
$ \min_{\theta} l(\theta) = \sum_{k=1}^{K} \frac{n_k}{n} l_k(\theta), \;\; \text{where} \;\; l_k(\theta)=\frac{1}{n_k} \sum_{i \in \mathbb{D}_k} l_i(\theta)$.
Note that, for machine learning problems, we usually consider $l_i(\theta) = l(y_i, f_\theta(x_i))$, where $l(\cdot,\cdot)$ is the loss of prediction on data point $(x_i, y_i)$ using the global model parameters $\theta$. At each time step $t$, the central server randomly selects a subset of clients ($S^t$) and exports the global weight vector $\theta^t$ to them. Each selected client $k \in S^t$ initializes its local model parameters with $\theta^t$, and then locally trains its own model using its own data $\mathbb{D}_k$ to obtain local update vector $\delta_k^{t+1}$. Finally each client $k$ sends $\delta_k^{t+1}$ to the server and then the server applies these updates to its global state to get $\theta^{t+1}$. This process repeats for several communication rounds until convergence of the central model.

As shown in Figure \ref{fig:motivation}, the local model updates sent by irrelevant clients (ones with bad data) can hamper stability as well as performance of the final centrally learned model. Hence, it is necessary to develop a mechanism that facilitates the server to prune irrelevant clients and not incorporate updates from them. In FL, the server has no visibility into how each client generates the updates. Hence, any mechanism that server uses to evaluate the relevance of a client has a limitation that it can only work with the updates it receives from clients treating each client as a black-box. 

\section*{Proposed Solution}
\begin{figure}
    \includegraphics[width=\linewidth]{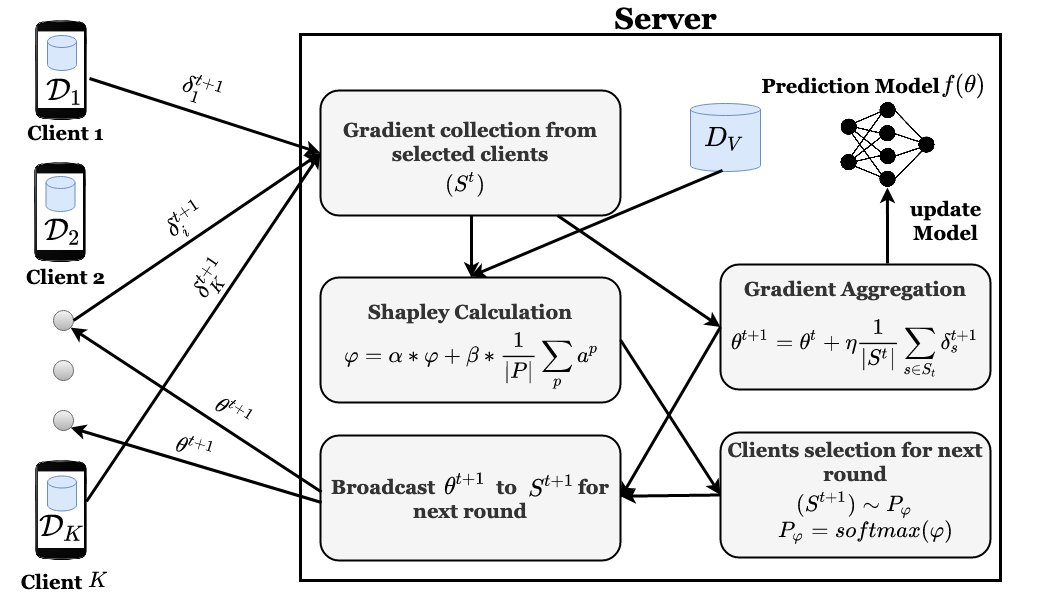}
    \caption{Steps that occur in communication round $t$ of Federated Relevant Client Selection (FRCS).}
    \label{fig:frcs_block_giagram}
\end{figure}

We propose an algorithm that models updates received from clients as players of a co-operative game. Then, we derive Shapley values for clients and use them to approximate the likelihood of a client being relevant. This paves way to an algorithm called S-FedAvg, which is further used to solve FRCS problems. Before delving into the solution approach, we briefly introduce some preliminary game-theoretic concepts here.

\subsection*{Preliminaries: Shapley Value}

{\bf Cooperative Games} \cite{myerson:1997,straffin:1993}: Let us now formalize the notions of a coalitional game and the Shapley value. To this end, we denote by $N = \{1,2,\ldots,n\}$ the set of players of a coalitional game. A \textit{characteristic function} $v: 2^N \to\mathbb{R}$ assigns to every coalition $C \subseteq N$ a real number representing payoff attainable by this coalition. By convention, it is assumed that $v(\emptyset)=0$. A \textit{characteristic function game} is then a tuple $(N,v)$.\\
{\bf Shapley Value:} It is usually assumed that the grand coalition, i.e., the coalition of all the agents in the game, forms. Given this, one of the fundamental questions of coalitional game theory is how to distribute the payoff of the grand coalition among the players. Among many different answers, Shapley \cite{Shapley1971} proposed to evaluate the role of each player in the game by considering its marginal contributions to all coalitions this player could possibly belong to. A certain weighted sum of such marginal contributions constitutes a player's payoff from the coalition game and is called the Shapley value \cite{myerson:1997,straffin:1993}. Let us see how to compute Shapley values.

Formally, let $\pi \in \Pi(N)$ denote a permutation of agents in $N$, and let $C_{\pi}(i)$ denote the coalition made of all predecessors of agent $i$ in $\pi$ (if we denote by $\pi(j)$ the location of $j$ in $\pi$, then: $C_{\pi}(i) = \{j \in \pi: \pi(j) < \pi(i)\}$). Then the Shapley value is defined as follows \cite{shapley:1996}:

\begin{equation}\label{originalShapley}
SV_i(v) = \frac {1}{|N|!} \sum_{\pi \in \Pi}[v(C_{\pi}(i) \cup \{i\}) - v(C_{\pi}(i))],
\end{equation}

i.e., the payoff assigned to $a_i$ in a coalitional game is the average marginal contribution of $a_i$ to coalition $C_{\pi}(i)$ over all $\pi \in \Pi$.  It is easy to show that the above  formula can be rewritten as:
\begin{equation}\label{Original_SV2nd_form}
SV_i(v) = \sum_{C \subseteq A \setminus \{i\}}\frac {|C|!(|N|-|C|-1)!} {|N|!} [v(C \cup \{i\}) - v(C)].
\end{equation}

\subsection*{Computing Relevance Scores of Clients ($\varphi$)}
\label{proposed_model}

In FL setting, in communication round $t$, each client $k \in S^t$ initializes its local model with parameters $\theta^t$ downloaded from the server and trains it using its private data $\mathbb{D}_k$. It runs Stochastic Gradient Descent algorithm for $E$ number of epochs to get the local parameters $\theta_k^{t+1}$. Then it derives an update $\delta_k^{t+1}$ by computing the difference of its local parameters with the global parameters as $\delta_k^{t+1} = \theta_k^{t+1} - \theta^t$ and shares it to the server. Now, the server computes importance of each client by making use of the following cooperative game and a validation dataset $D_V$. We define a cooperative game, $(\delta^{t+1},v)$, wherein the set of updates of individual clients $\delta^{t+1} = \{\delta_s^{t+1}\}_{s \in S^t}$ is the set of players and $v$ is a characteristic function that attaches a value for each $X \subseteq S^{t}$. We define the value of each $X \subseteq S^t$ as performance of the model built using updates only from $X$ on validation dataset $D_V$.
\begin{linenomath}
\begin{align*}
    \theta_X^{t+1} &= \theta^t + \frac{1}{|X|} \sum\limits_{s\in X}\delta_s^{t+1} \numberthis \\
    v(X, D_V) &= \mathcal{P}\big(f_{\theta_X^{t+1}}, D_V\big) \label{eq:char_perf} \numberthis
\end{align*}
\end{linenomath}


where $\mathcal{P}$ function denotes the performance (accuracy) of the central model with parameters $\theta_X^{t+1}$ on the validation data $D_V$.

Let $\varphi = (\varphi_1, \varphi_2, \ldots, \varphi_K)$ be the relevance vector wherein $\varphi_k$ represents the relevance of client $k$. $\varphi$ is private to the server. We posit that more the relevance value for a client, more is its likelihood to be relevant and thereby more is its contribution to the objective of central model.  The server initializes the relevance vector to be uniform $\varphi_k = \frac{1}{K};\; \forall k \in S$ at $t=0$. Then at each round $t$, we obtain the Shapley value for selected clients $S_t$ from the cooperative game $(\delta^{t+1}, v)$ mentioned above. Let us denote the computed Shapley values as $sv(s)$ for each $s \in S_t$. Now, the server updates the relevance vector $\varphi$ as follows.
\begin{linenomath}
\begin{align*}
    \varphi_s = \alpha * \varphi_s + \beta * sv(s);\;\forall s \in S_t \numberthis
\end{align*}
\end{linenomath}

We note that even though server updates the relevance vector for only a subset of clients at each round, the effect of partial updation is levied on all the $K$ clients. As we will see in the proposed S-FedAvg algorithm, we compute the relative relevance of each client my marginalizing $\varphi$ as $P_{\varphi} = softmax(\varphi)$, where $softmax(\cdot)$ is the traditional softmax function. By definition, $sv(\cdot)$ can be negative. In our definition of cooperative game, a client would receive a negative Shapley value, when the update it shares adversely affects the performance of the central model when used in coalition with updates shared by other clients.

\subsection*{Proposed Algorithm: Shapley Values Based FedAvg (S-FedAvg)}
Based on the above framework, we propse S-FedAvg algorithm which is outlined in Algorithm \ref{s-fedavg}.


\begin{algorithm}[!hbtp]
\SetAlgoLined
\SetKwInOut{Input}{input}
\SetKwInOut{Output}{output}
 \Input{ $K$: the number of clients, $S$: set of clients, $\theta$: parameters of central model, $D_V$: validation dataset at the server, $\{\mathbb{D}_k\}_{k=1}^K$ training dataset at each client $k \in S$, $m$: number of clients selected in each round, $B$: minibatch size, $\eta$: learning rate of central model, $\eta_i$: local learning rate at each client $i$, $E$: number of local epochs, $T$: max number of communication rounds, $R$: number of monte-carlo simulations, ($\alpha,\beta$): relevance scores update parameters;}
 \Output{$\;\theta^T$: parameters of the learned central model, $\varphi$: relevance value vector for clients} \vspace{0.25cm}
 $\varphi_k \gets \frac{1}{K}\;\; \forall k \in S$; \\
 initialize central model model parameters $\theta_1$ randomly;\\
 \BlankLine
 \ForEach{round $t = 1,2, \cdots$, $T$}{%
    \textbf{server executes:}\\
    $P_\varphi[k] = \frac{exp(\varphi_k)}{\sum \limits_{k \in S}exp(\varphi_k)} \;\; \forall k \in S$\;
    $S^t \gets$ sample $m$ clients $\sim P_{\varphi}$\;
    \BlankLine
     \textbf{clients ($S^t$) execute:}\;
     \ForEach{client $k \in S^t$}{
        $\theta_{local} \gets \theta^t$; \textit{// initialize local model}\\
        \ForEach{local epoch $e \in \{1,2,\cdots,E\}$}{
            $D_k \gets$ select a minibatch of size $B \subseteq \mathbb{D}_k$\;
            $\theta_{local} \gets \theta_{local} - \eta_k \nabla l_k(D_k, \theta_{local})$;
        }
        $\delta_k^{t+1} = \theta_{local} - \theta^t$;
     }
     \BlankLine
     \textbf{server executes:}\;
     $\delta^{t+1} \gets \{\delta_s^{t+1}\}_{s\in S^t}$; \textit{// collect client updates}\\
     $sv \;\; \Longleftarrow\;\;$ $sv$\_update($\theta^t,D_V,\delta^{t+1}, R)$\;
     $\varphi_k \gets \alpha * \varphi_k + \beta * sv[k]\;\; \forall k \in S^t$
     
    }
 
\caption{Shapley values based Federated Averaging (S-FedAvg)} 
\label{s-fedavg}
\end{algorithm}


\begin{algorithm}
\label{update-shap-alg}
\SetAlgoLined
\SetKwInOut{Input}{input}
\SetKwInOut{Output}{output}
 \Input{$\theta^t$: Parameters of the cenral model, $D_V$: Validation dataset, $\delta^{t+1}$:updates of selected clients, $R$: number of monte-carlo simulations}
 \Output{ $sv$: Shapley values for Clients that shared $\delta^{t+1}$} 
 \BlankLine
 $N\gets$ set of clients in $\delta^{t+1}$\;
 $P \gets$ set of $R$ permutations of $\delta^{t+1}$ \\
\ForEach{permutation $p \in P$}{
     $S_{p,i} = \{j | j \in N \land p(j) \leq i \}$ \\
     $a^{p}_{i} \gets v(\{S_{p,i}\cup i\}, D_V) - v(S_{p,i}, D_V)$ \\
     \tcc{$v(\cdot, \cdot)$ obtained as in Eqn: \ref{eq:char_perf}}
     $sv[i] \gets sv[i] + \frac{1}{|P|} * a_i^p$\;
    }
\caption{Compute Shapley values of ($\delta^{t+1}, v$)}
\label{shapley_value_convex_games}
\end{algorithm}


In the beginning, server initializes $\varphi$, the relevance scores of clients uniformly and the parameters of the central model $\theta_0$ randomly. Then, it runs the {\em outer for} loop in Algorithm \ref{s-fedavg} for $T$ communication rounds where in each round $t$, it does the following. First, the server computes a probability distribution $P_{\varphi}$ by applying softmax to the relevance vector $\varphi$. Then, it samples $m$ clients using $P_{\varphi}$ to form the sampled set of clients for receiving updates, call it $S_t$, for round $t$. In all our experiments, we set $T = 100$ and $m = 5$. Next, the server signals the clients in $S_t$ to share the updates. Each client on receiving the signal executes the {\em inner for} loop in Algorithm \ref{s-fedavg}. 

Each client $k \in S^t$ first downloads the model parameters $\theta^t$ and initialize their local model with it. We note that the architecture of local model is exactly same as that of the server. Then, it learns a local model by minimizing its local loss function $l_k$ using its local dataset $\mathbb{D}_k$. Clients use stochastic gradient descent algorithm to fit the local model for $E$ number of epochs with a local learning rate of $\eta_k$. In all our experiments, we set $\eta_k = 0.01$ and $E=5$. We further decay $\eta_k$ with a factor of $0.995$ every $20$ communication rounds at the server. Each selected client on fitting their local model derives the update using $\delta_k^{t+1} = \theta_{local} - \theta^t$ and shares it to the server.
Once, server receives updates from all the clients in $S^t$, it first computes the Shapley values according to Algorithm \ref{shapley_value_convex_games} using a co-operative game $(\delta^{t+1},v)$. Since computing Shapley values takes time complexity exponential in order of number of players, they are approximated by monte-carlo simulations in general \cite{castro-cor-2009,fatima-aij-2008,bachrach-aamas-2008}. We note that in all our experimental settings, we have $m=5$ players in the game. In such a setting, computing exact Shapley values requires computations over $m!=120$ permutations which is manageable. However, to test the robustness of the proposed approach, we set number of monte-carlo simulations as $R=10$ and run the {\em for loop} in Algorithm \ref{shapley_value_convex_games} for $10$ permutations only. For each permutation $p$ in the {\em for loop}, we define $S_{p,i} = \{j | j \in N \land p(j) \leq i \}$ (wherein $p(j)$ indicates the position of client $j$ in the permutation $p$). Then we compute its marginal value $a^{p}_{i} = v(\{S_{p,i}\cup i\}, D_V) - v(S_{p,i}, D_V)$. We then compute average of these marginal values and update the relevance scores according to the equation $\alpha * \varphi_k + \beta * sv[k]\;\; ; \forall k \in S^t; \alpha \in (0,1]$. In our experiments, we set $\alpha=0.75$ and $\beta=0.25$ respectively. Intuitively, $\alpha$ acts as the emphasis server places on past performance of clients and $\beta$ servers as the emphasis server places on current performance of clients respectively. So for a selected client $k \in S^t$, on expanding its relevance computation equation, we get

\begin{linenomath}
\begin{align*}
    \varphi_k &= \beta * sv^t[k] + \alpha * \mathcal{I}_{k\in S^{t-1}} * sv^{t-1}[k]  \\
    &+ \alpha^2 * \mathcal{I}_{k\in S^{t-2}} *  sv^{t-2}[k] \numberthis
\end{align*}
\end{linenomath}

where $\mathcal{I}_{k\in S^l}$ is the indicator that $k$ was sampled during $l^{th}$ communication round. $sv^{l}[k]$ here indicates the Shapley value of client $k$ at round $l$, which is undefined if $k$ if it is not sampled during $l$. We can see that for a client to have a good relevance score at time $t+1$, it should not only share relevant updates in the current round $t$, but also should have shared relevant updates in the earlier rounds $\{1,2,\cdots,t-1\}$ whenever it was sampled by the server. The entire training schedule of $FRCS$ is elucidated in the Figure \ref{fig:frcs_block_giagram}.
\section*{Experiments}

\subsection*{Datasets and Experimental Setup}
\label{sec:datasets_expts}
We describe the experimental setup for MNIST \footnote{MNIST: \url{http://yann.lecun.com/exdb/mnist/}} in detail here. In the interest of space, we provide the details about the experimental setup and results for other datasets in Supplementary material \footnote{Supplementary material: \url{https://github.com/nlokeshiisc/SFedAvg-AAAI21}}. The MNIST dataset is composed of handwritten digits. It has around $60000$ training samples and $10000$ test samples. Both the train and test samples are uniformly distributed across the $10$ labels. In all our experiments, we consider that $|S| = 10$ clients participate in a collaborative training process to build a classifier model. The central server samples $m = 5$ clients at each communication round and receives updates from them. Without loss of generality, we assume that the objective of the central server is to build a $5$ class classification neural network to classify images of even labels. $i.e.,$ the target classes are $\{0, 2, 4, 6, 8\}$. 

To simulate irrelevant clients, we inject Label-noise into the data possessed by them. In literature, Label-noise is further classified into open-set noise and closed-set noise \cite{openset_noise}. In closed-set label noise, data samples of a known class are mislabelled as data samples of a known class (eg., the image of a dog is mislabelled as a cat in dog vs. cat classification task). Whereas, in open-set label noise, data samples of an unknown class is assigned a label of a known class (eg., an image of a horse is mislabelled as a cat in cat vs. dog classification task). In this paper, we inject {\em open-set label noise} to irrelevant clients.

Henceforth, we refer to data corresponding to even digits as {\em relevant data} and those corresponding to odd digits as {\em irrelevant data}. As is typical with the federated learning setting, we assume that the data is distributed non-iid with each client. We follow the extreme {\em 1-class-non-iid} approach mentioned in \cite{federatedNonIID} while distributing the data to clients. Following their approach, we sort the data first according to their labels and then partitioned them and distributed evenly across the clients. The net effect is that each client would possess data of $1$ or $2$ consecutive labels only and hence their distributions are mostly not identical with each other.

We consider two types of experimental settings which we call {\em (a) relevant data setting} and {\em (b) irrelevant data setting}. In both the cases, server contains a dataset of about $5000$ even digits which is further partitioned into $D_V, D_{Test}$ such that $|D_V| = 1000$ and $|D_{Test}|=4000$ respectively. In $(a)$, we assume that all $10$ clients are relevant and thus possess images of even digits only. We assign all the even digits of $MNIST$ train dataset to the clients in the {\em 1-class-non-IID} manner mentioned above. The size of local dataset possessed by each client is $|\mathbb{D}_k| \approx \frac{30000}{10}$. In $(b)$, we assume that there are $6$ relevant and $4$ irrelevant clients. First, we distribute images of even digits to relevant clients. We sort the relevant data according to their labels and divide them into $6$ equi-sized partitions and distribute a partition to each client. To the remaining $4$ irrelevant clients, we assign images of odd digits as follows. First, we create a random bijective map from odd to even labels. Eg., one such map is $\kappa = \{1\rightarrow0, 5\rightarrow2\, 3\rightarrow4, 9\rightarrow6, 7\rightarrow8\}$. Then we flip label $O_l$ of each odd image $O$ as $label(O) = \kappa(O_l)$. After this step, all the images of odd digits would have even labels. We partition this noise-induced dataset among the $4$ irrelevant clients in a strategy similar to $(a)$.

{\bf Hyper-Parameter Values:} The values of hyper-parameters that we use in the experiments are:   $K=10, m=5, B=32, \alpha=0.75, \gamma=0.95, \lambda=2\%, \beta=0.25, \zeta=5, R=10, \eta_i = 0.01$.



The central server's objective is to prune away the updates from irrelevant clients and incorporate those from relevant clients in the model update process.
Further, to address the local minima problem, one can perhaps use pre-trained models like Imagenet, BERT etc., wherever appropriate. This in fact also would solve model initialization problem to a large extent.


Our proposed Algorithm \ref{s-fedavg} enables the server to compute the relevance values of clients which can then be normalized into a probability distribution to select irrelevant clients with high probability.

\subsection*{S-FedAvg: To Select Relevant Clients}
\begin{figure}[!hbtp]
    \centering
    \begin{subfigure}
    \centering
    \includegraphics[width=0.48\linewidth]{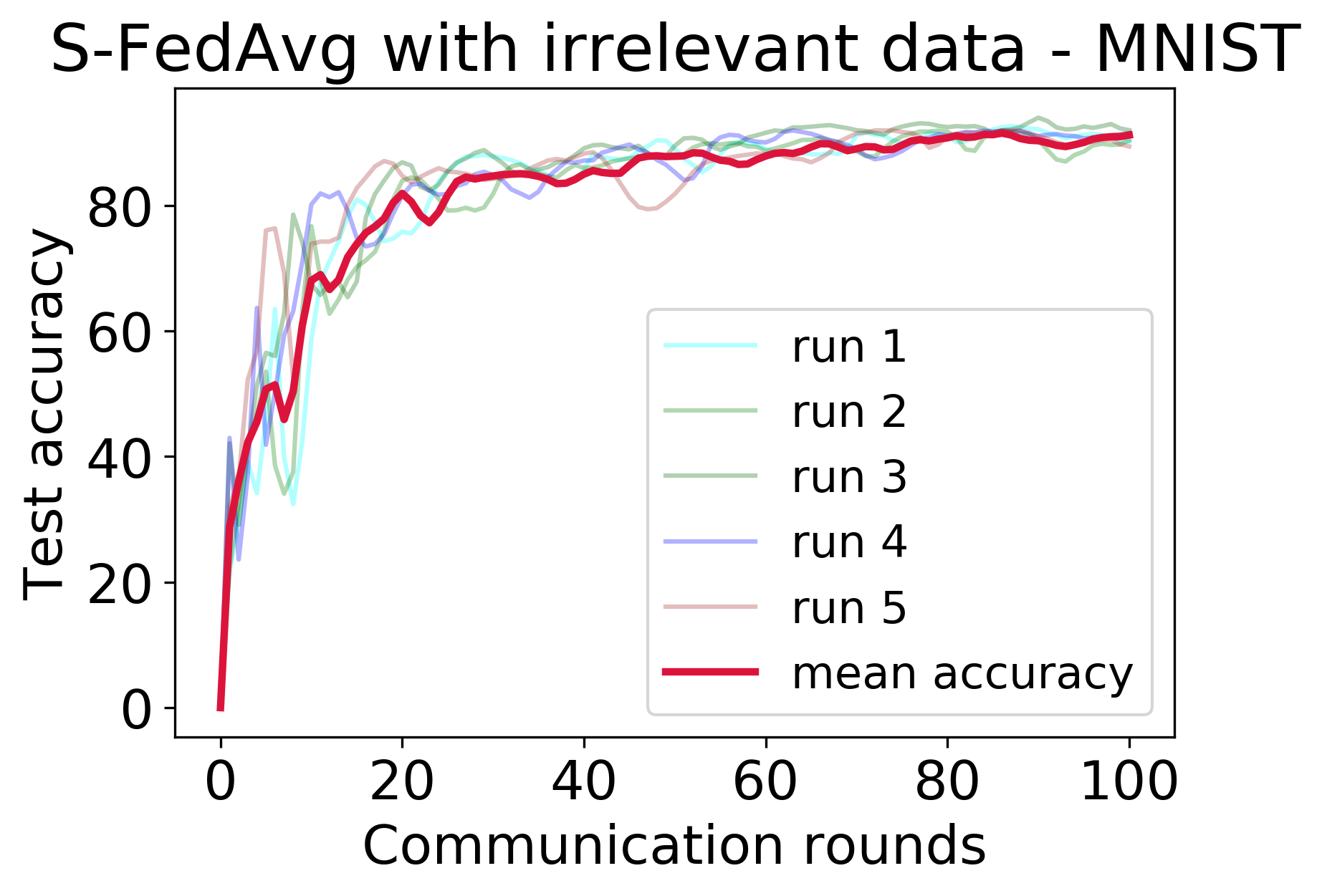}
    \end{subfigure}
    \begin{subfigure}
    \centering
    \includegraphics[width=0.48\linewidth]{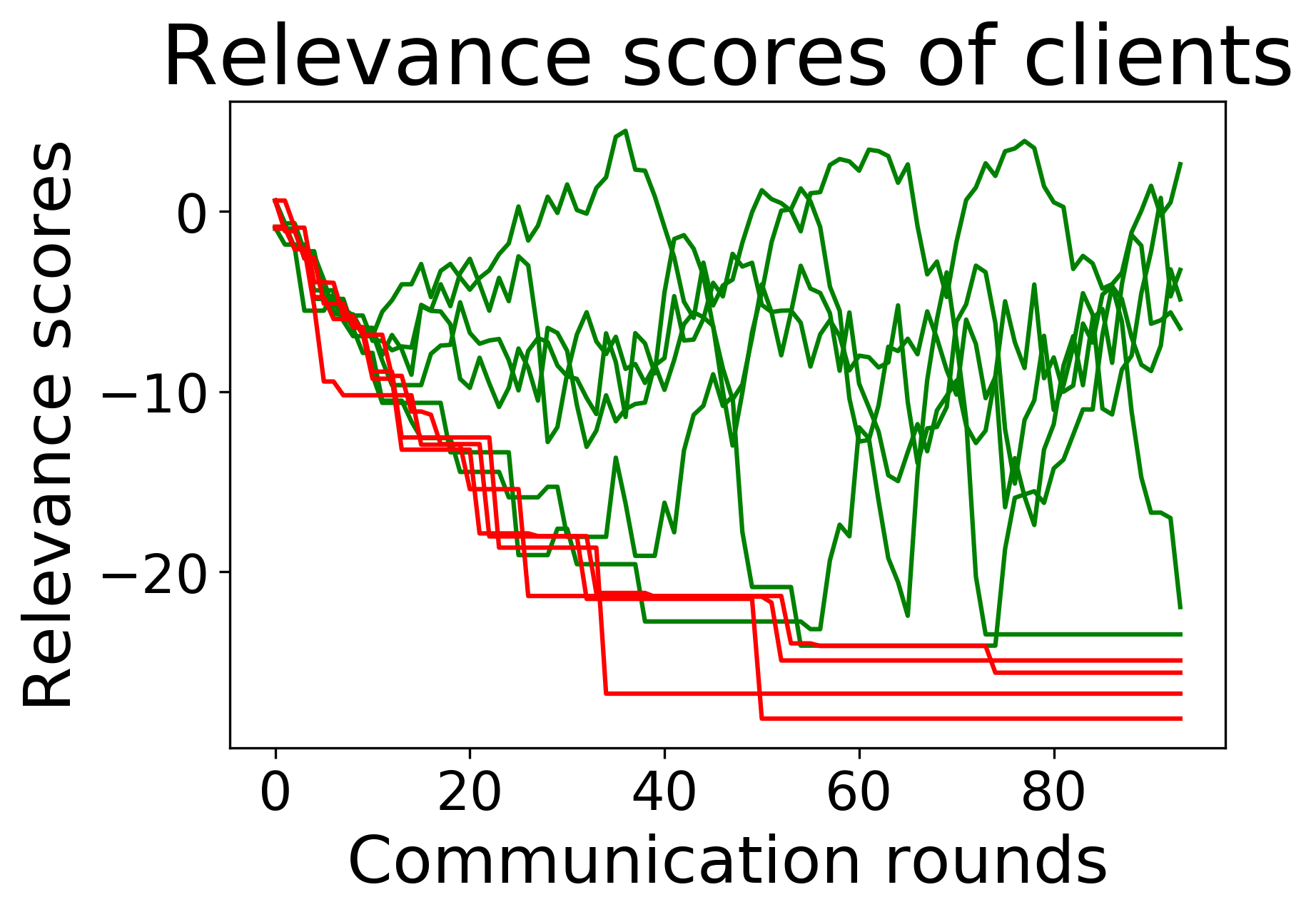}
\end{subfigure}
\caption{Performance of S-FedAvg algorithm with the irrelevant data setting.}
\label{fig:shap_avg}
\end{figure}

In this experiment, we show that $S-FedAvg$ algorithm assigns higher relevance scores to relevant clients. We apply S-FedAvg to {\em irrelevant data setting} case mentioned in \ref{sec:datasets_expts}. Recall that there are $6$ relevant clients and $4$ irrelevant clients in this case. Figure \ref{fig:shap_avg}(a) shows the accuracy of the central model learned using our algorithm across $100$ communication rounds. We can observe that the central model converges smoothly here which is completely in contrast to how a central model trained with FedAvg performs as shown in Figure \ref{fig:motivation}. To explain the supremacy in performance, we look at Figure \ref{fig:shap_avg}(b) that shows the relevance scores $\varphi$ of clients across $100$ rounds. The green lines there show the relevance scores of $6$ relevant clients and the red lines show the relevance scores of the $4$ irrelevant clients respectively. We can easily see that relevance scores of relevant clients are consistently higher than their irrelevant counterparts. Even though the relevance score fluctuates up and down, we can consistently see the trend that relevant clients dominate irrelevant clients. 


\subsection*{Class-Specific Best Client Selection}
\label{class_specific_best_client_selection}

While Algorithm \ref{s-fedavg} helps in identifying the relevant clients with respect to central model's objective, an equally important problem at the server is to identify class-specific best clients.
This problem is important particularly in several domains such as healthcare and retail, wherein data imbalance problem is prevalent and the server may want to incentivize the clients that possess data belonging to the minority class.

\begin{figure}[!hbtp]
    \centering
    \begin{subfigure}
    \centering
    \includegraphics[width=0.48\linewidth]{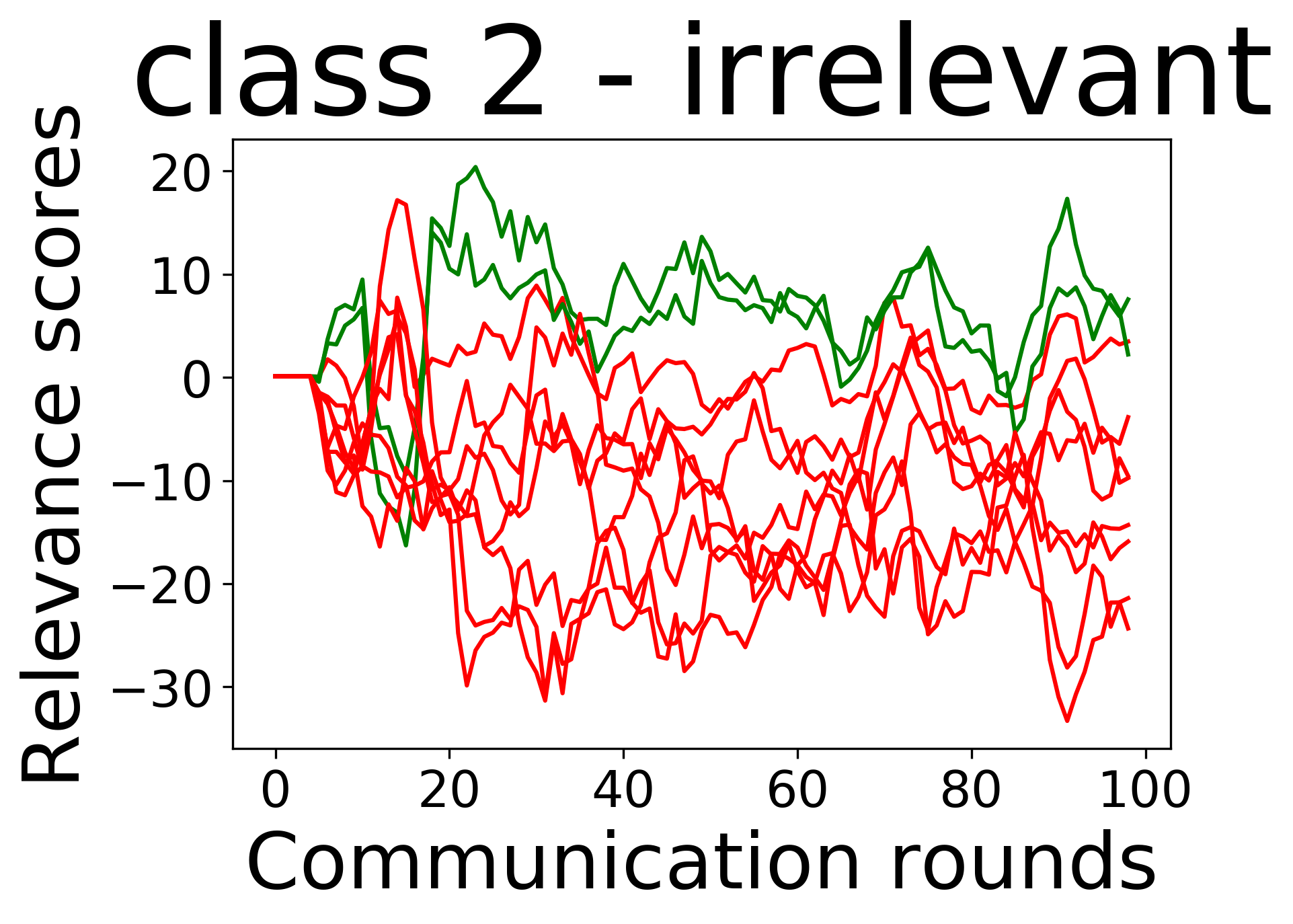}
    \end{subfigure}
    \begin{subfigure}
    \centering
    \includegraphics[width=0.48\linewidth]{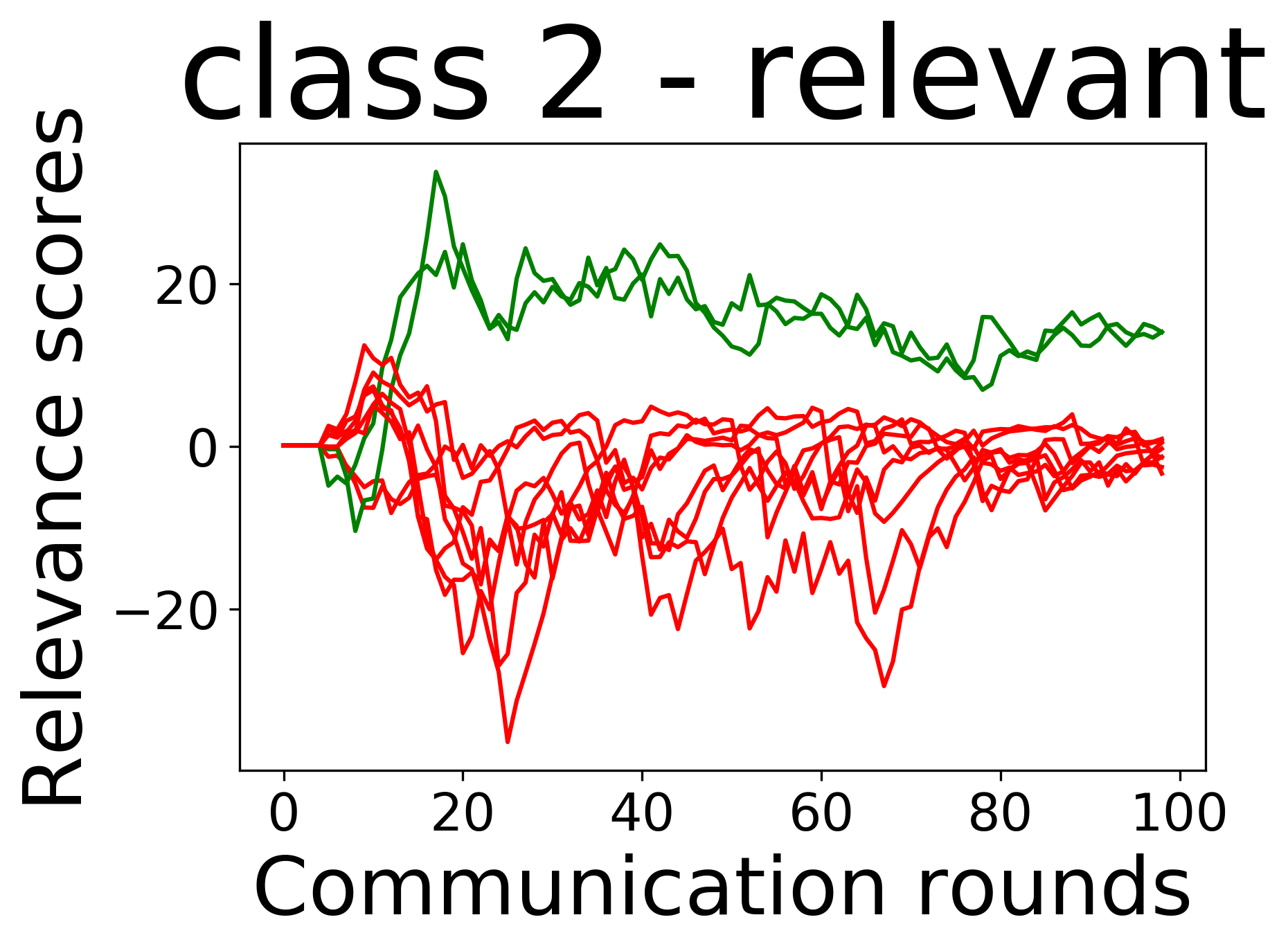}
\end{subfigure}
\caption{Relevance value $\varphi_c$ for class $C=2$.}
\label{fig:cls_specific_bad_good_nodes}
\end{figure}

We tackle this problem as follows:\\
(1) Consider that we want to identify clients that possess best data w.r.t. class label $c$. We sample all the data points of class $c$ from the validation data $D_V$ at the server and we refer to this as $V_c$. 
(2) While running S-FedAvg algorithm, we present a method to learn the class specific relevance values, which we call $(\varphi_c)$, for each client at each communication round $t$. To learn $\varphi_c$, we invoke Algorithm \ref{shapley_value_convex_games} by setting $D_V = V_c$. 
(3) Because the validation set only comprises of the data instances of class $c$, the clients that possess data whose updates when shared would help in accurately classifying test points of class $c$ would receive higher relevance values in $\varphi_c$. We run the experiments to detect the class-specific best nodes for each $c \in \{0, 2, 4, 6, 8\}$. The results are shown in the Figure  \ref{fig:cls_specific_bad_good_nodes}. In the interest of space, we show only for class $2$. In each of these plots, the green curves correspond to relevance values of the clients with data of class $2$. It is easy to see from the plots that the clients that possess data of label $2$ do have relatively higher relevance values than the ones without data of label $2$. This separation is more explicit in the relevant data setting as the gap between the green and red curves is more significant. Results for other classes is also similar and are available in supplementary material.


\subsection*{Data Label Standardization}
\label{sec:data_lbl_std}
Consider the scenario where a client might possess relevant data but their labels are inconsistent w.r.t. the labeling convention of the server. That is, what is label $x$ according to the server could be label $y$ according to the client due to some ambiguity. One such case where ambiguity could arise is: server assumes that the first neuron of the output layer of the central model corresponds to label $x$, whereas the client believes that the first neuron of the output layer corresponds to label $y$ and $x\ne y$. 
We refer to this problem as {\em data label standardization} or {\em label permutation problem} wherein the clients have to discover the corrupted data samples then permute their labels back to the correct labels. Hereafter, we call a client whose labels are permuted as {\em corrupted client}.


Since corrupted clients adversely affect the model, S-FedAvg algorithm duly would assign low relevance values to such clients and thus the probability of such clients getting sampled by the server would be less. However, the updates of corrupted clients would be useful to the server if their accurate labels are recovered. To this end, we propose an algorithm called S-FedAvg-Label-Std that detects each corrupted client and then determines the correct label permutation. This algorithm works as follows:

(1) We run S-FedAvg Algorithm \ref{s-fedavg} for some initial communication rounds until the model stabilizes. We call a model stabilized when its performance on $D_V$ doesn't fluctuate $\lambda \%$ for at least $\zeta$ consecutive rounds. 

(2) Once the model stabilizes, the central server signals the possible corrupted clients and asks them to inspect for label permutation. As part of this signal, the server also sends a performance dictionary  $P_{dict} = \{label : accuracy\}$ to each client. $P_{dict}$ contains class-specific accuracies (also called class wise precision) of the central model in that communication round. These performances are again computed using $D_V$. 
(3) Each signaled client $k$ downloads both the central model's parameters ($\theta^{t}$) and the performance dictionary $P_{dict}$ from the server. Then client $k$ executes the following steps: 
(3.a) Let the local data owned by client $k$ have $q$ labels $\{l_1, l_2, \ldots, l_q\}$. These $q$ labels are a subset of target labels. Client $k$ splits $\mathbb{D}_k$ label-wise into $q$ partitions to get $\mathbb{D}_k = \{D^k_{l_1}, D^k_{l_2}, \ldots, D^k_{l_q}\}$ respectively. Let $n^k_{l_i} = |D^k_{l_i}|$ be the size of the partition corresponding to the label $l_i$.

(3.b) Client $k$ does the following for each partition in $D^k_{l_i}$. For each of the $|n^k_{l_i}|$ points belonging to the partition, the client gets the predictions from its local model initialized with $\theta^t$. Let $L$ be the label predicted for a majority of data points in the partition. Let $n_L$ be the number of points whose labels are predicted as $L$ by the model. We note that $L$ need not necessarily belong to one of the $q$ labels that the client possesses. Now, the client computes the fraction $n_L / n^k_{l_i} * 100$ and if this value is more than the accuracy corresponding to label $L$ in the performance dictionary, i.e. $P_{dict}(L)$, the client $k$ swaps its local label $l_i$ with label $L$ for all the data points in $D^k_{l_i}$. This is why we need a stabilized model otherwise $P_{dict}$ computed from a unstable may suggest incorrect permutation to the clients.

(3.c) Each client that does a permutation acknowledges the server that it has rectified the label permutation issue. 

\begin{figure}[t]
    \centering
    \begin{subfigure}
    \centering
    \includegraphics[width=0.48\linewidth]{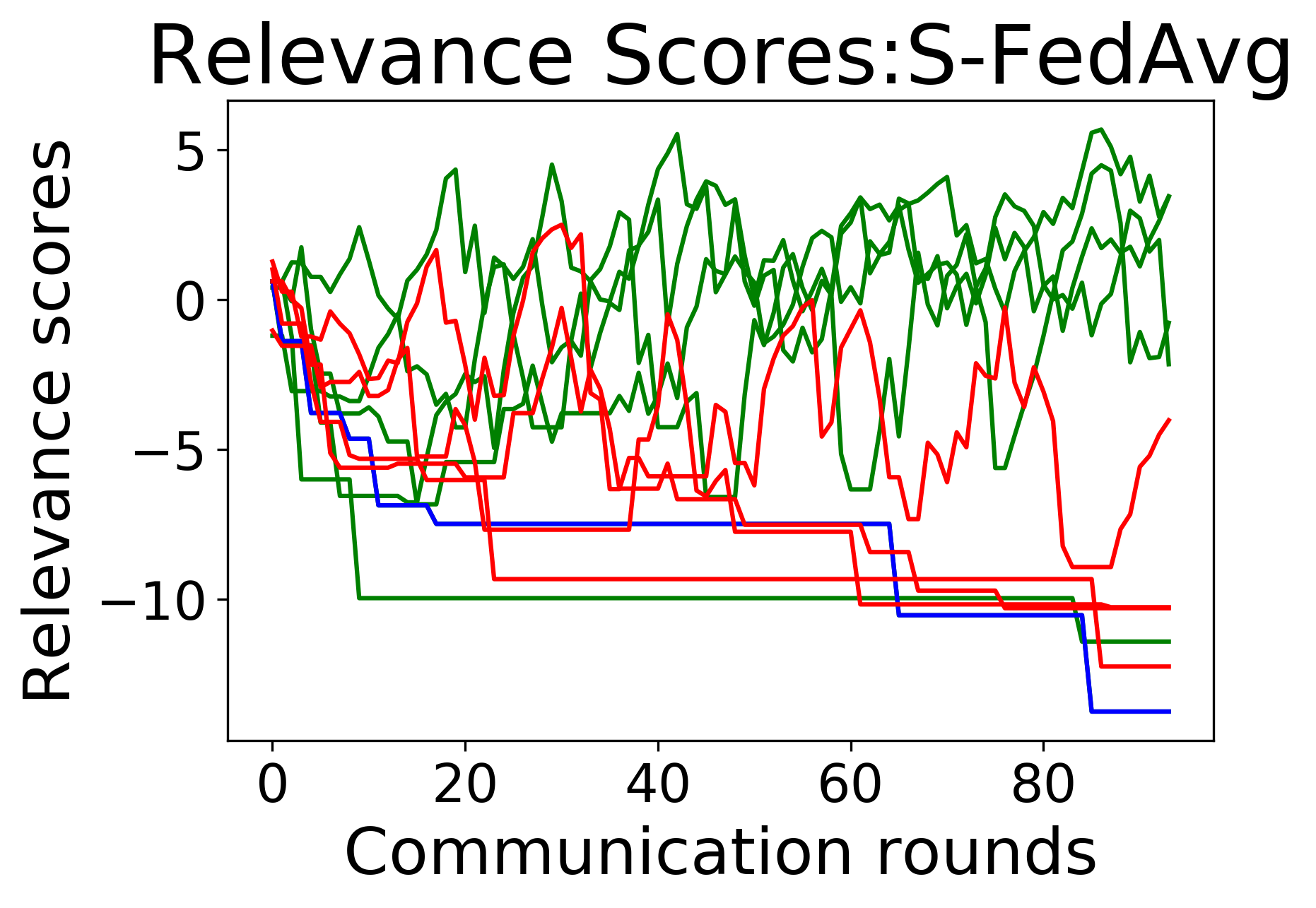}
    \end{subfigure}
    \begin{subfigure}
    \centering
    \includegraphics[width=0.48\linewidth]{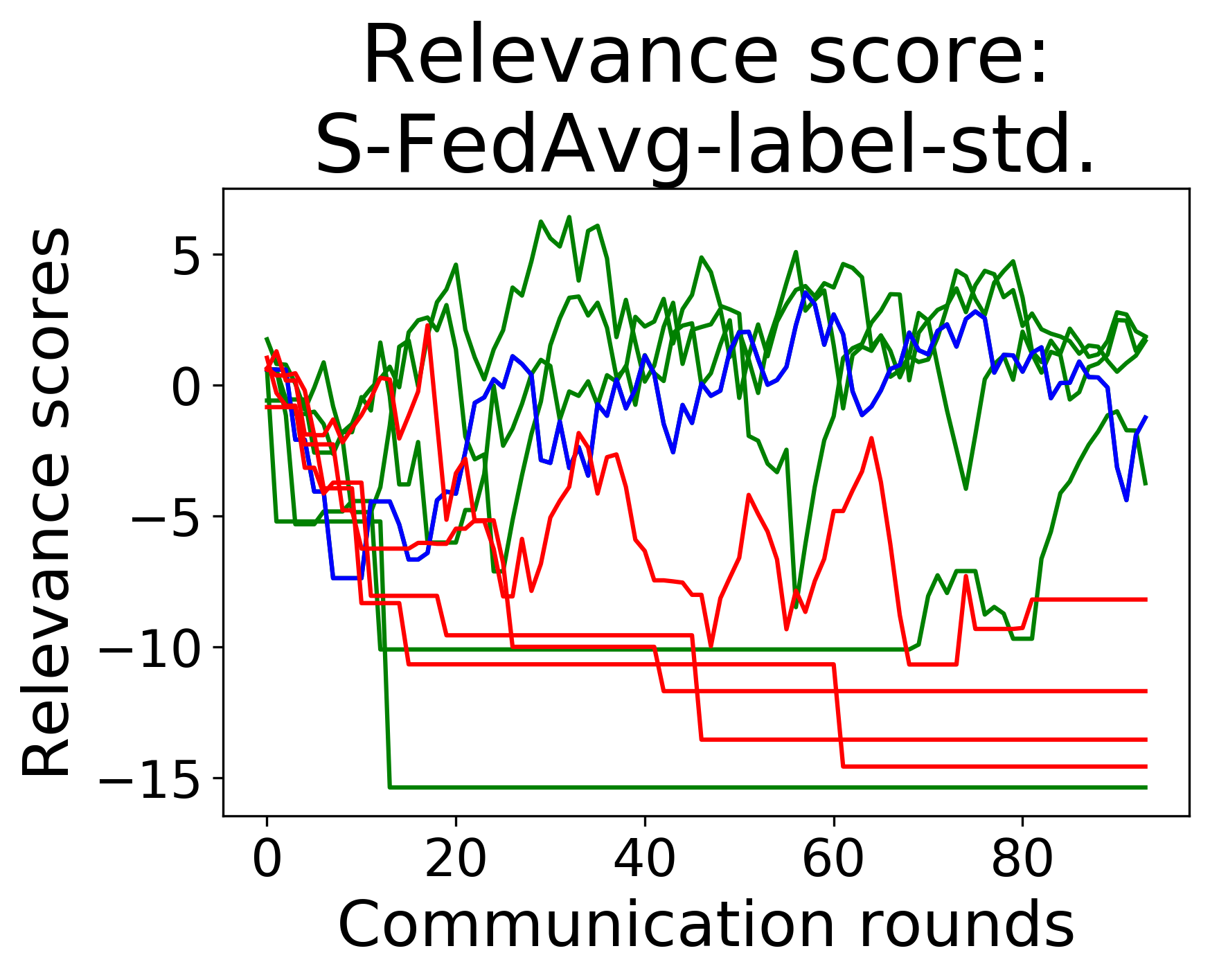}
    \end{subfigure}
    
    \centering
    \begin{subfigure}
    \centering
    \includegraphics[width=0.48\linewidth]{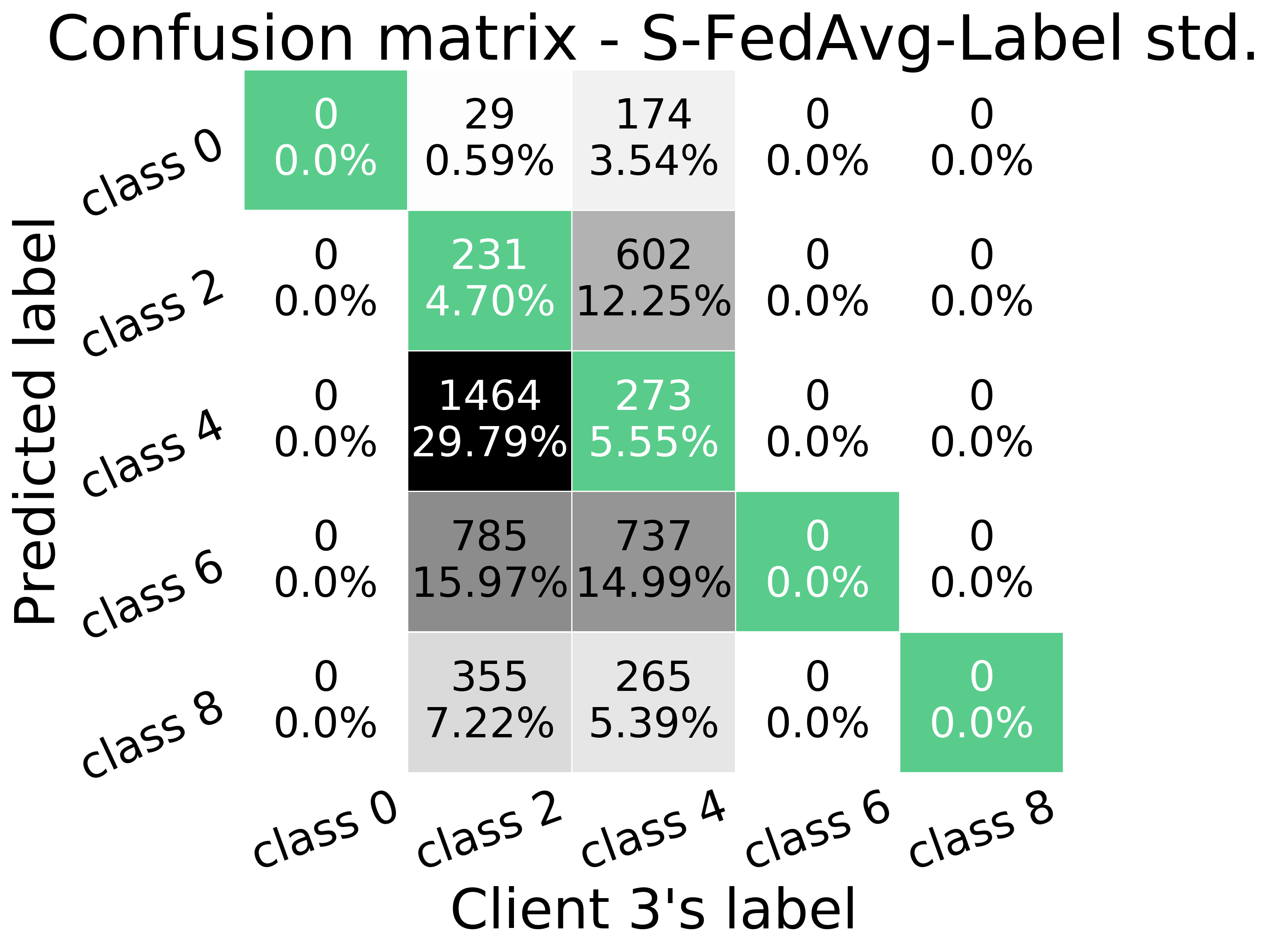}
    \end{subfigure}
    \begin{subfigure}
    \centering
    \includegraphics[width=0.48\linewidth]{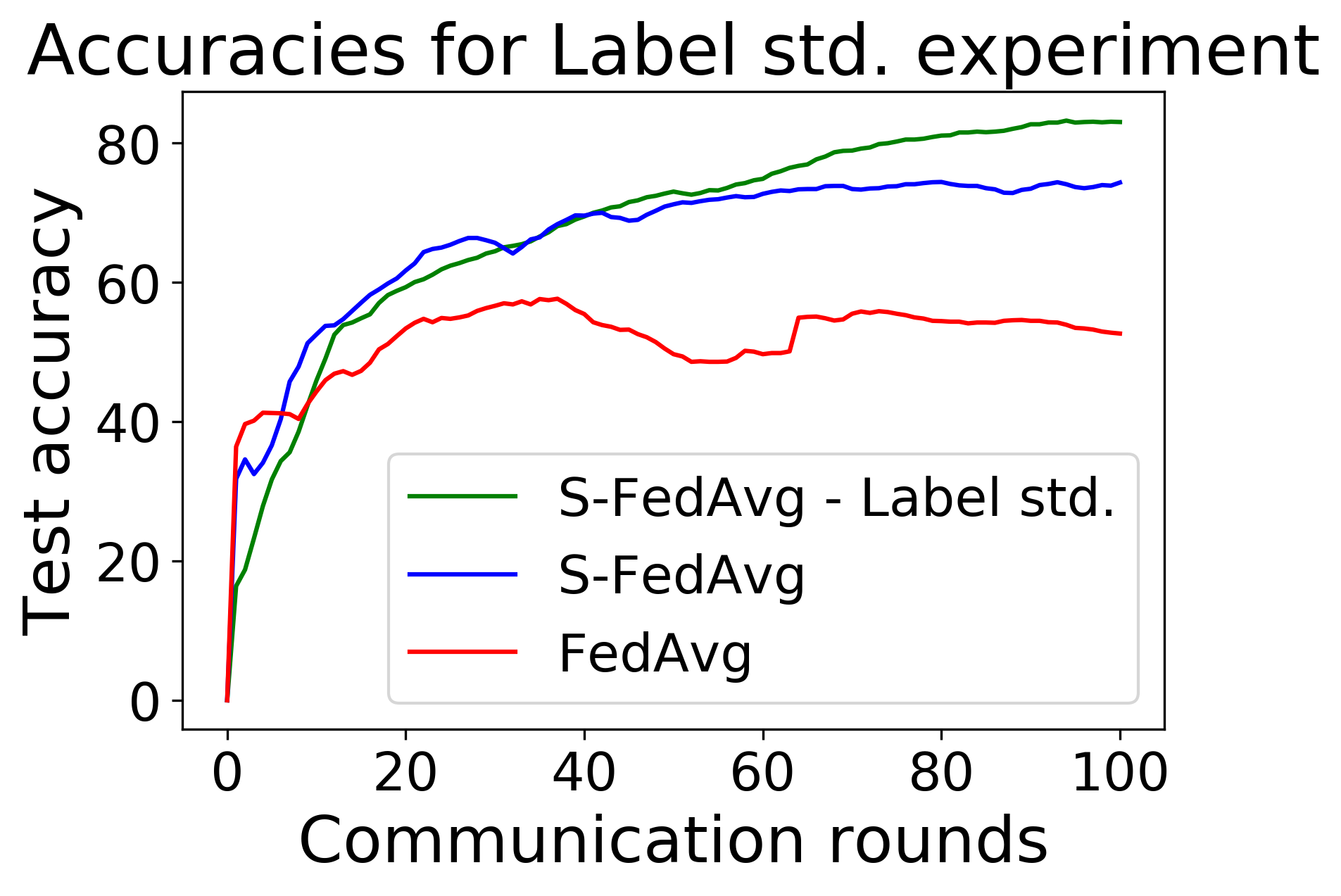}
    \end{subfigure}
\caption{(a) S-FedAvg-label-std (b)Experimental results for Data label standardization experiment}
\label{fig:data_lbl_std_1_2}
\end{figure}

On receiving the acknowledgement, if the server believes that client may start contributing positively, it subdues the penalty levied in relevance score so far. One such way of doing so is: server adjusts the relevance scores of these clients to the mean of the relevance vector $mean(\varphi[l])$.
This step makes sure that heavy penalties levied by the S-FedAvg algorithm on corrupted are curtailed so that their chances of getting sampled (after label correction) by the central server would increase in the subsequent rounds. The results for data label standardization are shown in Figure \ref{fig:data_lbl_std_1_2}.

We consider the irrelevant data case and manually corrupt the labels of client $3$. We swap its label $2$ with $4$. Figures \ref{fig:data_lbl_std_1_2} (top) show the relevance values of clients across communication rounds using S-FedAvg and S-FedAvg-Label-Std respectively. The green curves correspond to relevant clients and red curves correspond to irrelevant clients respectively. Here, the relevance value of the corrupted client is denoted by the blue curve. From Figure \ref{fig:data_lbl_std_1_2} (top-right), we observe that label permutation occurred at communication round $17$. As expected, the relevance value for the corrupted client (i.e. client $3$) is heavily penalized by S-FedAvg. Figure \ref{fig:data_lbl_std_1_2} also shows the confusion matrix of predictions obtained from the downloaded server model using parameters $\theta_{17}$ at communication round $t=17$.




As we can see in the column corresponding to label $2$, predictions from the model are in favor of label $4$ and hence the corrupted client swaps label $2$ with label $4$. Note that after permutation, data with label $2$ may still be inconsistent and if they are, they hopefully will get fixed in the subsequent rounds. Figure \ref{fig:data_lbl_std_1_2} (bottom-right) shows the test performance of three algorithms viz., FedAvg, S-FedAvg, and S-FedAvg-Label-Std.
The performance of S-FedAvg-Label-Std is superior to that of S-FedAvg and FedAvg; and this performance gain is due to our proposed data label standardization procedure.

\section*{Conclusions}
We have shown in this paper that presence of even a very less number of irrelevant clients could adversely affect the performance of the federated learning model. To this end, we developed a systematic method to tackle this problem and its variants. The following are three interesting directions to future work. First, development of a scalable approximation method to compute Shapley values to deal with FL settings with
thousands of clients. 
Second, instead of using Shapley values to select clients as is done in this paper, its interesting to investigate the impact on the model if we use Shapley values as the 
averaging weights in model aggregation. 
Third, it would be interesting to consider cleaning the gradients from irrelevant clients and then incorporate them into the model update process rather than discarding them \cite{yu2020gradient}.

\section*{Broader Impact}
Here we present a brief discussion on the {\em potential broader impact} of our proposed methods in this paper. Our algorithms would significantly aid the federated learning techniques to succeed despite the fact that a few participating clients possess bad quality data. 
Since federated learning provides a natural framework for technologists to build cross-enterprise and hybrid cloud based AI solutions, our algorithms are of paramount importance for the success of such solutions. 
Our algorithms could be used as is in certain domains such as the healthcare industry. For example, consider a scenario wherein a set of hospitals are interested to collaboratively train a model for drug discovery purpose (for instance, this is very much timely scenario for COVID-19). Clearly, since the clinical data owned by each hospital is private and confidential as it contains sensitive patient information, federated learning framework is a natural solution approach to train the model. Further, since the clinical data owned by each hospital cannot be subjected to the central server's scrutiny, it is possible that such clinical data may be noisy and thus irrelevant. In this scenario, our proposed algorithms would be of significant value to build a better model.


\bibliography{AAAI}

\begin{thebibliography}{32}
\providecommand{\natexlab}[1]{#1}
\providecommand{\url}[1]{\texttt{#1}}
\providecommand{\urlprefix}{URL }
\expandafter\ifx\csname urlstyle\endcsname\relax
  \providecommand{\doi}[1]{doi:\discretionary{}{}{}#1}\else
  \providecommand{\doi}{doi:\discretionary{}{}{}\begingroup
  \urlstyle{rm}\Url}\fi

\bibitem[{Abadi et~al.(2016)Abadi, Chu, Goodfellow, McMahan, Mironov, Talwar,
  and Zhang}]{privacy1}
Abadi, M.; Chu, A.; Goodfellow, I.; McMahan, H.~B.; Mironov, I.; Talwar, K.;
  and Zhang, L. 2016.
\newblock Deep learning with differential privacy.
\newblock In \emph{Proceedings of the 2016 ACM SIGSAC Conference on Computer
  and Communications Security}, 308--318.

\bibitem[{Bachrach et~al.(2008)Bachrach, Markakis, Procaccia, Rosenschein, and
  Saberi}]{bachrach-aamas-2008}
Bachrach, Y.; Markakis, E.; Procaccia, A.; Rosenschein, J.; and Saberi, A.
  2008.
\newblock Approximating power indices.
\newblock In \emph{Proceedings of the 7th InternationalJoint Conference on
  Autonomous Agents and Multi-Agent Systems (AAMAS)}, 943--950.

\bibitem[{Barroso et~al.(2020)Barroso, Mart{\'i}nez-C{\'a}mara, Luz{\'o}n,
  Gonz{\'a}lez-Seco, Veganzones, and Herrera}]{client_selection_1}
Barroso, N.~R.; Mart{\'i}nez-C{\'a}mara, E.; Luz{\'o}n, M.; Gonz{\'a}lez-Seco,
  G.; Veganzones, M.~A.; and Herrera, F. 2020.
\newblock Dynamic Federated Learning Model for Identifying Adversarial Clients.
\newblock \emph{ArXiv} abs/2007.15030.

\bibitem[{Bergman and Hoshen(2020)}]{outlier_detection}
Bergman, L.; and Hoshen, Y. 2020.
\newblock Classification-Based Anomaly Detection for General Data.
\newblock \emph{ArXiv} abs/2005.02359.

\bibitem[{Blanchard et~al.(2017)Blanchard, El~Mhamdi, Guerraoui, and
  Stainer}]{client_selection_2}
Blanchard, P.; El~Mhamdi, E.~M.; Guerraoui, R.; and Stainer, J. 2017.
\newblock Machine Learning with Adversaries: Byzantine Tolerant Gradient
  Descent.
\newblock In Guyon, I.; Luxburg, U.~V.; Bengio, S.; Wallach, H.; Fergus, R.;
  Vishwanathan, S.; and Garnett, R., eds., \emph{Advances in Neural Information
  Processing Systems 30}, 119--129. Curran Associates, Inc.
\newblock
  \urlprefix\url{http://papers.nips.cc/paper/6617-machine-learning-with-adversaries-byzantine-tolerant-gradient-descent.pdf}.

\bibitem[{Bonawitz et~al.(2019)Bonawitz, Eichner, Grieskamp, Huba, Ingerman,
  Ivanov, Kiddon, Konecn{\'y}, Mazzocchi, McMahan, Overveldt, Petrou, Ramage,
  and Roselander}]{federated_scale}
Bonawitz, K.; Eichner, H.; Grieskamp, W.; Huba, D.; Ingerman, A.; Ivanov, V.;
  Kiddon, C.; Konecn{\'y}, J.; Mazzocchi, S.; McMahan, H.; Overveldt, T.~V.;
  Petrou, D.; Ramage, D.; and Roselander, J. 2019.
\newblock Towards Federated Learning at Scale: System Design.
\newblock \emph{ArXiv} abs/1902.01046.

\bibitem[{Choi et~al.(2020)Choi, Hong, Lee, and Lim}]{choicenet}
Choi, S.; Hong, S.; Lee, K.; and Lim, S. 2020.
\newblock Task Agnostic Robust Learning on Corrupt Outputs by
  Correlation-Guided Mixture Density Networks.
\newblock \emph{2020 IEEE/CVF Conference on Computer Vision and Pattern
  Recognition (CVPR)} 3871--3880.

\bibitem[{Fatima, Wooldridge, and Jennings(2008)}]{fatima-aij-2008}
Fatima, S.; Wooldridge, M.; and Jennings, N. 2008.
\newblock A linear approximation method forthe shapley value.
\newblock \emph{Artificial Intellilgence} 172(14): 1673--1699.

\bibitem[{Geyer, Klein, and Nabi(2017)}]{privacy2}
Geyer, R.~C.; Klein, T.; and Nabi, M. 2017.
\newblock Differentially private federated learning: A client level
  perspective.
\newblock \emph{arXiv preprint arXiv:1712.07557} .

\bibitem[{Ghorbani and Zou(2019)}]{datashapley}
Ghorbani, A.; and Zou, J. 2019.
\newblock Data shapley: Equitable valuation of data for machine learning.
\newblock \emph{arXiv preprint arXiv:1904.02868} .

\bibitem[{Haddadpour and Mahdavi(2019)}]{sgd_fed_convergence}
Haddadpour, F.; and Mahdavi, M. 2019.
\newblock On the Convergence of Local Descent Methods in Federated Learning.
\newblock \emph{ArXiv} abs/1910.14425.

\bibitem[{J.~Castro(2009)}]{castro-cor-2009}
J.~Castro, D.~Gomez, J.~T. 2009.
\newblock Polynomial calculation of the shapley valuebased on sampling.
\newblock \emph{Computers \& Operations Research} 36(5): 1726--1730.

\bibitem[{Kang et~al.(2019)Kang, Xiong, Niyato, Yu, Liang, and
  Kim}]{client_selection_5}
Kang, J.; Xiong, Z.; Niyato, D.; Yu, H.; Liang, Y.-C.; and Kim, D.~I. 2019.
\newblock Incentive design for efficient federated learning in mobile networks:
  A contract theory approach.
\newblock In \emph{2019 IEEE VTS Asia Pacific Wireless Communications Symposium
  (APWCS)}, 1--5. IEEE.

\bibitem[{Kim et~al.(2018)Kim, Park, Bennis, and Kim}]{privacy3}
Kim, H.; Park, J.; Bennis, M.; and Kim, S.-L. 2018.
\newblock On-device federated learning via blockchain and its latency analysis.
\newblock \emph{arXiv preprint arXiv:1808.03949} .

\bibitem[{Li et~al.(2019)Li, Huang, Yang, Wang, and Zhang}]{li2019convergence}
Li, X.; Huang, K.; Yang, W.; Wang, S.; and Zhang, Z. 2019.
\newblock On the convergence of fedavg on non-iid data.
\newblock \emph{arXiv preprint arXiv:1907.02189} .

\bibitem[{McMahan et~al.(2017{\natexlab{a}})McMahan, Moore, Ramage, Hampson,
  and y~Arcas}]{communication_efficiency}
McMahan, B.; Moore, E.; Ramage, D.; Hampson, S.; and y~Arcas, B.~A.
  2017{\natexlab{a}}.
\newblock Communication-efficient learning of deep networks from decentralized
  data.
\newblock In \emph{Artificial Intelligence and Statistics}, 1273--1282. PMLR.

\bibitem[{McMahan et~al.(2017{\natexlab{b}})McMahan, Moore, Ramage, Hampson,
  and y~Arcas}]{mcmahan:2017}
McMahan, H.; Moore, E.; Ramage, D.; Hampson, S.; and y~Arcas, B.~A.
  2017{\natexlab{b}}.
\newblock Communication-Efficient Learning of Deep Networks from Decentralized
  Data.
\newblock In \emph{In Proceedings of the 20th International Conference on
  Artificial Intelligence and Statistics (AISTATS)}, 1273--1282.

\bibitem[{Monderer and Shapley(1996)}]{shapley:1996}
Monderer, D.; and Shapley, L. 1996.
\newblock Potential Games.
\newblock \emph{Games and Economic Behavior} 14: 124--143.

\bibitem[{Myerson(1997)}]{myerson:1997}
Myerson, R.~B. 1997.
\newblock \emph{Game Theory: Analysis of Conflict}.
\newblock Cambridge, Massachusetts, USA: Harvard University Press.

\bibitem[{Patrini et~al.(2017)Patrini, Rozza, Menon, Nock, and
  Qu}]{label_noise_approach_1}
Patrini, G.; Rozza, A.; Menon, A.; Nock, R.; and Qu, L. 2017.
\newblock Making Deep Neural Networks Robust to Label Noise: A Loss Correction
  Approach.
\newblock 2233--2241.
\newblock \doi{10.1109/CVPR.2017.240}.

\bibitem[{Recht et~al.(2011)Recht, Re, Wright, and Hogwild}]{recht-nips-2011}
Recht, B.; Re, C.; Wright, S.; and Hogwild, F.~N. 2011.
\newblock A lock-free approach to parallelizing stochastic gradient descent.
\newblock In \emph{Proceedings of Advances in Neural Information Processing
  Systems (NIPS)}, 693--701.

\bibitem[{Shapley(1971)}]{Shapley1971}
Shapley, L. 1971.
\newblock Cores of Convex Games.
\newblock \emph{Int J of Game Theory} 1: 11--26.

\bibitem[{Smith et~al.(2018)Smith, Forte, Chenxin, Takác, Jordan, , and
  Jaggi}]{smith-jmlr-2018}
Smith, V.; Forte, S.; Chenxin, M.; Takác, M.; Jordan, M.~I.; ; and Jaggi, M.
  2018.
\newblock Cocoa: A general framework for communication-efficient distributed
  optimization.
\newblock \emph{Journal of Machine Learning Research} 18.

\bibitem[{Song, Tong, and Wei(2019)}]{client_selection_4}
Song, T.; Tong, Y.; and Wei, S. 2019.
\newblock Profit Allocation for Federated Learning.
\newblock In \emph{2019 IEEE International Conference on Big Data (Big Data)},
  2577--2586. IEEE.

\bibitem[{Straffin(1993)}]{straffin:1993}
Straffin, P.~D. 1993.
\newblock \emph{Game Theory and Strategy}, volume~36.
\newblock Mathematical Association of America, 1 edition.
\newblock ISBN 9780883856376.
\newblock \urlprefix\url{http://www.jstor.org/stable/10.4169/j.ctt19b9kx1}.

\bibitem[{{Verma}, {White}, and {de Mel}(2019)}]{dinesh-verma-2019}
{Verma}, D.; {White}, G.; and {de Mel}, G. 2019.
\newblock Federated AI for the Enterprise: A Web Services Based Implementation.
\newblock In \emph{2019 IEEE International Conference on Web Services (ICWS)},
  20--27.
\newblock \doi{10.1109/ICWS.2019.00016}.

\bibitem[{Wang et~al.(2018)Wang, Liu, Ma, Bailey, Zha, Song, and
  Xia}]{openset_noise}
Wang, Y.; Liu, W.; Ma, X.; Bailey, J.; Zha, H.; Song, L.; and Xia, S.-T. 2018.
\newblock Iterative learning with open-set noisy labels.
\newblock In \emph{Proceedings of the IEEE Conference on Computer Vision and
  Pattern Recognition}, 8688--8696.

\bibitem[{Yang et~al.(2019)Yang, Liu, Chen, and Tong}]{yang-survey-2019}
Yang, Q.; Liu, Y.; Chen, T.; and Tong, Y. 2019.
\newblock Federated Machine Learning: Concept and Applications.
\newblock \emph{ACM Trans. Intell. Syst. Technol} 10(2).

\bibitem[{Yu et~al.(2020)Yu, Kumar, Gupta, Levine, Hausman, and
  Finn}]{yu2020gradient}
Yu, T.; Kumar, S.; Gupta, A.; Levine, S.; Hausman, K.; and Finn, C. 2020.
\newblock Gradient Surgery for Multi-Task Learning.
\newblock In Larochelle, H.; Ranzato, M.; Hadsell, R.; Balcan, M.~F.; and Lin,
  H., eds., \emph{Advances in Neural Information Processing Systems},
  volume~33, 5824--5836. Curran Associates, Inc.
\newblock
  \urlprefix\url{https://proceedings.neurips.cc/paper/2020/file/3fe78a8acf5fda99de95303940a2420c-Paper.pdf}.

\bibitem[{Zhan et~al.(2020)Zhan, Li, Qu, Zeng, and Guo}]{client_selection_3}
Zhan, Y.; Li, P.; Qu, Z.; Zeng, D.; and Guo, S. 2020.
\newblock A learning-based incentive mechanism for federated learning.
\newblock \emph{IEEE Internet of Things Journal} .

\bibitem[{Zhao et~al.(2018)Zhao, Li, Lai, Suda, Civin, and
  Chandra}]{federatedNonIID}
Zhao, Y.; Li, M.; Lai, L.; Suda, N.; Civin, D.; and Chandra, V. 2018.
\newblock Federated Learning with Non-IID Data.
\newblock \emph{ArXiv} abs/1806.00582.

\bibitem[{Zhu, Liu, and Han(2019)}]{deepleekage}
Zhu, L.; Liu, Z.; and Han, S. 2019.
\newblock Deep leakage from gradients.
\newblock In \emph{Advances in Neural Information Processing Systems},
  14774--14784.

\end{thebibliography}
\end{document}